\pdfoutput=1
\documentclass[runningheads]{llncs}
\usepackage{graphicx}
\usepackage{algorithmic}
\usepackage{graphicx}
\usepackage{textcomp}
\usepackage{xcolor}
\usepackage{url}
\usepackage{multirow}
\usepackage{amsfonts,amsmath}
\usepackage{enumitem,kantlipsum}
\usepackage{booktabs}
\usepackage{bbm}
\usepackage{lipsum}
\usepackage{wrapfig}
\usepackage[bottom]{footmisc}
\usepackage{hyperref}
\usepackage{subfigure}
\usepackage{diagbox}
\newcommand{\bx}{\mathbf{x}}
\newcommand{\bz}{\mathbf{z}}
\newcommand{\bs}{\mathbf{s}}
\newcommand{\bq}{\mathbf{q}}

\newcommand{\bh}{\mathbf{h}}

\newcommand{\CE}{\text{E}}

\newcommand{\BR}{\mathbb{R}}
\newcommand{\BN}{\mathbb{N}}

\newcommand{\BW}{\mathbf{W}}

\newcommand{\ind}{\mathbbm{1}}

\DeclareMathOperator*{\argmax}{\arg\!\max}

\begin{document}
\title{Option Tracing: Beyond Correctness Analysis \\ in Knowledge Tracing}
\author{Aritra Ghosh$^1$, Jay Raspat$^2$, Andrew Lan$^1$}
\authorrunning{A. Ghosh et al.}
\institute{$^1$University of Massachusetts Amherst\footnote{This work is supported by the National Science Foundation under grant IIS-1917713. We also thank the reviewers for their constructive feedback.}, $^2$ Independent Consultant}
\maketitle              %

\begin{abstract}
Knowledge tracing refers to a family of methods that estimate each student's knowledge component/skill mastery level from their past responses to questions. 
One key limitation of most existing knowledge tracing methods is that they can only estimate an \emph{overall} knowledge level of a student per knowledge component/skill since they analyze only the (usually binary-valued) correctness of student responses. Therefore, it is hard to use them to diagnose specific student errors. 
In this paper, we extend existing knowledge tracing methods beyond correctness prediction to the task of predicting the exact option students select in multiple choice questions. 
We quantitatively evaluate the performance of our option tracing methods on two large-scale student response datasets. We also qualitatively evaluate their ability in identifying common student errors in the form of clusters of incorrect options across different questions that correspond to the same error. 
\end{abstract}

\section{Introduction}

Knowledge tracing (KT) \cite{kt} refers to a family of student modeling methods that estimate student mastery levels on knowledge components/skills/concepts from their past responses to questions/items and predict their future performance. These estimates and predictions can be used to i) provide feedback to students on their progress, especially in intelligent tutoring systems \cite{bevbook} and ii) drive personalization, i.e., selecting the action that each learner should take next to maximize their learning outcomes \cite{rlreview,dash,ritter}. Many different KT methods have been developed, ranging from hidden Markov model-based Bayesian knowledge tracing methods \cite{mozerfuse,ktcomparepardos,yudelson}, factor analysis-based methods such as learning factor analysis \cite{lfa}, performance factor analysis \cite{pfa}, and the item Difficulty, student ability, skill, and student skill practice history (DAS3H) method \cite{das3h}, to deep learning-based methods \cite{akt,sakt,rkt,dkt,gikt,dkvmn}. These methods have enjoyed various degrees of success; some of these methods, including most Bayesian knowledge tracing and factor analysis-based methods, exhibit excellent interpretability while other, deep learning-based methods trade off interpretability for excellent predictive accuracy on students' future performance. 

However, one key limitation of these KT methods is that they operate exclusively on (usually binary-valued)
response data that indicates whether a student responds to a question correctly or not. 
Therefore, they can only estimate students' \emph{overall} mastery level on each knowledge component. 
However, not all incorrect responses are equal: there can be numerous incorrect ways to answer a math question \cite{mlp}, caused by different underlying errors. Studies have shown that only a fraction of incorrect answers generated by students can be anticipated and explained by cognitive models integrated into intelligent tutoring systems \cite{koedinger,ritter,kurt}, teachers \cite{neilmath}, and numerical simulations \cite{neilmath,selent}. 
Typical underlying errors include having a ``buggy rule'' \cite{brown}, exhibiting a certain misconception \cite{feldman,junchen,smith}, or a general lack of knowledge on certain knowledge components \cite{anderson}. 
Since it is hard to diagnose such student errors from correctness data alone, we need to develop KT methods that analyze full student responses.  

\begin{figure}[tp]

    \centering
    \includegraphics[width=0.8\columnwidth]{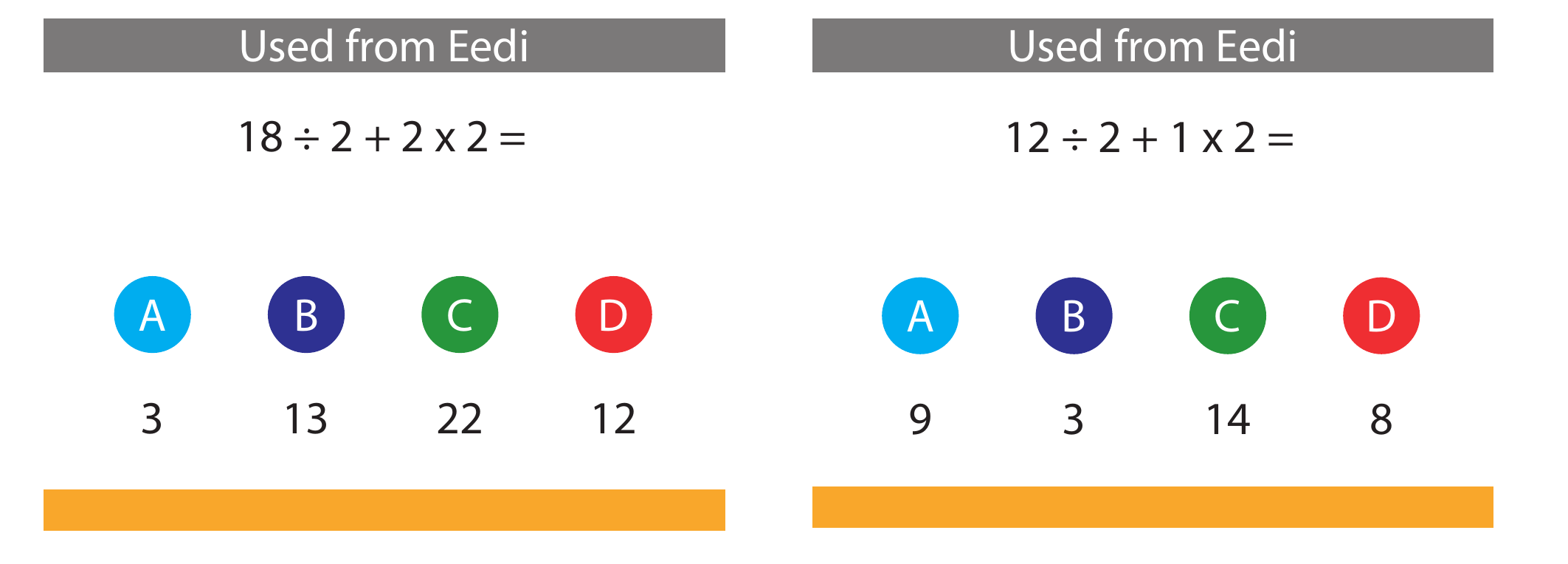}
    \vspace{-0.2cm}
    \caption{Some distractor options in well-designed MCQs are potentially capable of capturing typical student errors. Option C in both questions here correspond to the error of not mastering the order of operations and always working left to right.}
    \label{fig:misc}
    \vspace{-0.5cm}
\end{figure}

Some datasets, including the large-scale  Eedi\footnote{\href{https://eedi.com/projects/neurips-education-challenge}{https://eedi.com/projects/neurips-education-challenge}} \cite{neuripschal} and EdNet\footnote{\href{https://github.com/riiid/ednet}{https://github.com/riiid/ednet}} \cite{ednet} datasets, contain the exact options students select on multiple choice questions (MCQs); this option data provides us with an opportunity to extend existing KT methods to analyze specific student option selections rather than their answer correctness. 
In an ideal situation, well-designed MCQs should have well-crafted incorrect distractor options that each corresponds to one or more typical student errors; 
Figure~\ref{fig:misc} shows an example from the Eedi dataset for two questions on the subject brackets, indices, division, multiplication, addition, subtraction (BIDMAS). Option C in both questions correspond to the same error of not fully mastering ``order of operations'' and always working left to right. 
However, manually identifying these errors is an unscalable and labor-intensive process since most existing MCQs do not come with consistent labels on the error(s) underlying each incorrect option. 
Therefore, it is important to explore whether we can develop KT methods to identify errors each incorrect option corresponds to and potentially diagnose student errors automatically. 
These methods would then be useful through i) informing teachers to communicate with students to understand the source of their errors, ii) enabling the development of automated feedback \cite{vt}, and iii) enabling the design of alternative instructional approaches such as asking students to criticize erroneous examples \cite{errorex}.

\subsection{Contributions}
In this paper, we develop option tracing (OT), a KT framework that uses the exact option each student selects on each question as both input and predicted output. We extend several existing KT methods to the OT setting, including a long short-term memory (LSTM) network-based method, deep knowledge tracing (DKT) \cite{dkt}, 
a graph convolutional network-based method, graph-based interaction model for knowledge tracing (GIKT) \cite{gikt}, 
and an attention network-based method, attentive knowledge tracing (AKT) \cite{akt}. We emphasize that the goal of this paper is \textbf{NOT} to compare all KT methods; instead, our goal is to study how can we generalize them to analyze student option selections in MCQs. Therefore, we only study some representative methods.
We conduct the following experiments on the Eedi and EdNet datasets:
First, we quantitatively evaluate our OT methods under both the collaborative filtering (CF) setup (introduced by the NeurIPS 2020 Education Challenge \cite{neuripschal}) and the typical KT setup on the task of option prediction. 
Second, we qualitatively demonstrate the interpretability exhibited by our OT framework using clustering algorithms to group incorrect options across multiple questions into clusters of shared underlying errors. Results show that the learned clusters match up with those manually identified by a domain expert to some degree. Therefore, OT can potentially offer a \emph{bottom-up} approach for error identification by extracting student errors from actual data instead of the typical \emph{top-down} approach of anticipating errors before seeing data. Our implementation will be publicly available at \url{https://github.com/arghosh/OptionTracing}.

\section{Related Work}
\label{sec:rw}

The options students select in MCQs can be regarded as a type of \emph{categorical data}, which has previously been studied in both the item response theory (IRT) and recommender systems research communities. However, in both cases, most prior works focus on the case where the categories are \emph{ordered}. In IRT research, polytomous IRT-based models \cite{mbitmc,sparfatag,polytomous} are used to model students' responses with multiple ordered categories, such as letter grades and partial credits. In recommender systems research, neural collaborative filtering (NCF)-based methods are used to model star ratings provided by users on items \cite{ncf}. 
There are relatively few models for \emph{unordered} categorical data such as the nominal response model (NRM) from the IRT research community, which has been applied to the analysis of MCQs \cite{nrm,ncd}.
\vspace{-0.25cm}
\section{Data and Problem Setup}
\label{sec:dataps}

The Eedi dataset contains the responses of more than 100,000 students to 27,613 MCQs across $389$ labeled subjects, totaling over $15$ million responses over the course of more than a year. Each response corresponds to the exact option a student selected on each question (among four options, $\{$A,B,C,D$\}$). 
We will also use a small subset of the Eedi data where we have access to the exact question (in the form of images) for quantitative analysis; this dataset contains the responses of more than $4,900$ students to $948$ questions, totaling in over $1.3$ million responses. 
The EdNet dataset contains the responses of more than 700,000 students to 13,169 MCQs across $189$ labeled subjects, totaling over $95$ million responses over the course of more than two years.

We use two experimental setups for evaluation purposes.
First, in the CF setup, the task is to predict each student's responses to a subset of questions that they responded to, given their responses to other questions (possibly in the future). Popular methods for this setup are neural collaborative filtering (NCF) \cite{ncf} and graph convolutional networks (GCN) \cite{gcmc,gcn}. 
Second, in the KT setup for evaluating KT methods, the task is to predict each student's responses to future questions based on their entire past response history.

\vspace{-0.25cm}
\subsection{Problem Setup}

Each student's performance record consists of a sequence of responses to questions assigned at a series of discrete time steps. 
For student $i$ at time step $t$, we denote the combination of the question that they answered, the set of subjects this question covers, their binary-valued response correctness, the option they chose, and the correct option to this question as a tuple, $(q_t^i, \{s_{t,j}^i\}_{j=1}^{n_{t}^i}, r_t^i, y_t^i, c_t^i)$, where $q_t^i\in \BN^{+}$ is the question index, $s_{t,j}^i\in \BN^{+}$ denotes the index of the  $j^{\text{th}}$ subject, $j\in 1, \ldots, n_t^i$ since each question can be tagged with multiple subjects, $r_t^i\in \{0,1\}$ is the response correctness ($1$ corresponds to a correct response),
$y_{t}^i\in \{A,B,C,D\}$ is the option the student selected, and $c_{t}^i\in \{A,B,C,D\}$ is the correct option for this question. 
In the CF setup, we associate a mask variable $m_t^i\in \{0,1\}$ with each time step, where $1$ represents that the timestep is part of the training set. This variable helps us to mask out responses we need to predict when we compute the training loss. Given observed responses $\{(q_t^i, \{s_{t,j}^i\}_{j=1}^{n_{t}^i},  r_t^i, y_t^i, c_t^i)\}_{t:m_t^i=1}$, the task is to predict the exact options students select on questions in the test set, i.e., $y_{t'}^{i'}$ for $(t',i'): m_t^i=0$. In the KT setup, we observe each student's entire history of responses to questions; thus, given their past history up to time $t\!-\!1$ as $\{(q_{\tau}^i, \{s_{{\tau},j}^i\}_{j=1}^{n_{{\tau}}^i}, r_{\tau}^i, y_{\tau}^i, c_{\tau}^i)\}_{\tau=1}^{t-1}$, our goal is to predict $y_t^i$ at the current time step, $t$. Under these notations, existing KT methods focus on predicting response correctness, $r_t^i$.

\vspace{-0.25cm}
\section{Methodology}
In this section, we detail our OT methods for both the CF and KT setups. 
Before delving into the individual methods, we start with a set of unified modules that apply to all methods in this paper. The question embedding module $\CE_{q}:q\rightarrow \BR^d$ transforms the question index $q_t^i$ to a $d$-dimensional, learnable real-valued vector in $\BR^d$. Similarly, the response embedding module $\CE_{r}:r\rightarrow \BR^d$ transforms the response correctness $r_t^i$ to $\BR^d$ and the option embedding module $\CE_o: \{A,B,C,D\}\rightarrow \BR^d$ transforms the correct option $c_t^i$ and the chosen option $y_t^i$ to vectors in $\BR^d$. We do not use separate embeddings for every question-option $(q,o)$ pair since that leads to overfitting in our experiments; instead, the $2d$-dimensional embedding for $(q,o)$ is obtained using $[\CE_q(q)\oplus \CE_o(o)]$ where $\oplus$ is the concatenation operator. The subject embedding module $\CE_{s}:s\rightarrow \BR^d$ transforms the subject index to $\BR^d$. Since each question may be tagged with several subjects, we define the final subject embedding as $\CE_s(\{s_{t,j}^i\}_{j=1}^{n^i_t}) =  \sum_{j=1}^{n_t^i} \CE_s(s_{t,j}^i)$.  Some of the methods (such as NCF) use a user embedding module $\CE_{u}:i\rightarrow \BR^d$ that transforms the student index to $\BR^d$. For simplicity, we use the same $d$-dimensional vector for all embedding modules; however, the dimensions of each module can be different.  We train all  model parameters, denoted as $\Theta$, which contains the embeddings listed here and other model parameters specific to each individual method, by minimizing the negative log-likelihood of the selected options as
\begin{align*}
  \underset{\Theta}{\text{minimize}}\quad -\textstyle\sum_{i=1}^{|\text{Students}|}\textstyle\sum_{t=1}^{|\text{Sequence}_i|}\textstyle\sum_{o\in \{A,B,C,D\}}\ind[y_t^i=o]\log p(o |q_t^i ;\Theta),
\end{align*}
where $\ind$ is the indicator function.
Since the options are unordered categories, the resulting loss function corresponds to the common cross-entropy loss \cite{dlbook}. 

\vspace{-0.25cm}
\subsection{Option Prediction under the CF Setup}
\label{sec:cfmodel}
\textbf{NCF.} 
NCF is one of the most popular CF methods for user-item interaction data. In the option prediction task, students correspond to users and questions corresponds to items. The input for NCF at time step $t$ for student $i$, $\bx_t^i$, is
\[\bx_t^i= [\CE_q(q_t^i)\, \oplus\, \CE_u(i)\, \oplus\, \CE_s(\{s_{t}^i\})].\]
Predictive probabilities $p(y_t^i = o)$ over four options $o\in \{A,B,C,D\}$ are calculated using the softmax function \cite{dlbook},
\begin{eqnarray*}
\bz_t^i = f(\bx_t^i)\in \BR^4,\quad p(y_t^i = o|\bx_t^i)= [\text{softmax}(\bz_t^i)]_{o},\quad
    \hat{y}_t^i = \argmax_{o \in \{A,B,C,D\}} [\bz_t^i]_{o},
\end{eqnarray*}
where $f(\cdot)$ denotes a feed-forward, fully-connected neural network and $[]_{o}$ refers to the $o^\text{th}$ entry of a vector. In NCF, the model parameters are the weights and biases in the feed-forward neural network $f(\cdot)$; this prediction module is shared by the subsequent methods. 

\textbf{PO-BiDKT.}
The main drawback of NCF is that the student embedding is static and not updated as students answer more questions and their knowledge states evolve. 
Recurrent neural networks, and in particular LSTM-type models are capable of modeling evolving knowledge as hidden states \cite{dkt}. However, we cannot directly use methods such as DKT in the CF setup since the student's responses at some time steps in their response sequence are not observed. Therefore, we use the following method to handle evolving knowledge states using recurrent networks with missing observations. 
The input at each time step is given by
\begin{align}
\label{eq:input}
    \bx_t = [\CE_q(q_t^i) \,\oplus\, \CE_o(c_t^i) \, \oplus\,  \CE_s(\{s_{t}^i\}) \, \oplus\,  \big(\CE_o(y_t^i)\odot m_t^i\big) \, \oplus\,  \big(\CE_l(r_t^i)\odot m_t^i\big)],
\end{align}
where $\odot$ denotes the element-wise multiplication between two vectors. We mask the option embeddings and response correctness embeddings using $m_t^i$ for time steps where we do not observe them but still use the question embedding as input. 
We also extend the base LSTM module in DKT to a bi-directional LSTM (Bi-LSTM) \cite{graves}. 
Here, we compute two latent knowledge states using two separate LSTM modules, the forward state  $\overrightarrow{\bh}_t$ that summarizes the student's past response history and the backward state $\overleftarrow{\bh}_t$ that summarizes the student's future response history at time step $t$ as
$$\overrightarrow{\bh}_{t+1}= \text{Forward LSTM}(\overrightarrow{\bh}_t, \bx_t), \ \overleftarrow{\bh}_{t-1}= \text{Backward LSTM}(\overleftarrow{\bh}_t, \bx_t).$$
The final latent knowledge state is the concatenation of the two states as $\bh_t=[\overrightarrow{\bh}_t\oplus \overleftarrow{\bh}_t]$. The parameters include two sets of parameters for the forward and backward LSTMs in addition to the parameters for the fully connected network $f(\cdot)$. We call this method partially observed bi-directional DKT, or PO-BiDKT. 
The output to the prediction module is computed using
\begin{align}
\label{eq:output}
\bz_t^i = f([\bh_t^i \oplus \CE_q(q_t^i) \oplus \CE_o(c_t^i) \oplus \CE_s(\{s_t^i\}) ])\in \BR^4. 
\end{align}

\textbf{GCN-augmented PO-BiDKT (BiGIKT).} In our datasets, each question is tagged with a few subjects by question designers or domain experts. These subject tags provide important information on how these questions are related since we  expect questions from the same subject to have some shared features. GCNs excel at formulating these relations and learning from  graph-structured data. Since we can represent the question-subject association matrix using a bipartite graph, (loosely) following GIKT \cite{gikt}, we connect GCNs with PO-BiDKT to jointly learn question and subject embeddings using the structure imposed by the subject tags. In this method, we use hierarchical representations of subjects and questions: starting with initial subject and question embeddings $\CE_s(s_t^i)$ and $\CE_q(q_t^i)$, the first layer GCN embedding for the $j^{\text{th}}$ subject and the second layer GCN embedding for the $i^{\text{th}}$ question are computed as
\begin{eqnarray*}
  \bs_j^{1}\! =\! \text{tanh}\! \Big(\!\BW_s^s \CE_s(s_j)\!+\! \frac{\!\sum_{i \in N_j^s}\!\BW_s^q \CE_q(q_i)}{|N_j^s|}\!\Big)\!,\,\bq_i^{2}\! =\! \text{tanh}\!\Big(\!\BW_q^q \CE_q(q_i)\!+\! \frac{\sum_{j \in N_i^q}\!\BW_q^s \bs_i^{1}}{|N_i^q|}\!\Big), 
\end{eqnarray*}
where $N_j^s\,(N_i^q)$ denotes the set of questions (subjects) associated with subject (question) $s_j\,(q_i)$ and $\BW_s^s$, $\BW_s^q$, $\BW_q^s$ and $\BW_q^q$ are learnable parameter matrices. The hyperbolic tangent (tanh) non-linearity operate entry-wise on vectors.
We replace the subject embeddings $\CE_s(s_t^i)$ and the question embedding $\CE_q(q_t^i)$ in the base Bi-LSTM (Eq.~\ref{eq:input} and Eq.~\ref{eq:output}) with these GCN-based embeddings. The model parameters of this method include the GCN weight parameter matrices in addition to the Bi-LSTM parameters.

\subsection{Option Prediction under the KT Setup}
In the KT setup, we predict future responses using only past responses and assume that every past student response is observed. We extend several existing neural network-based KT methods for the option prediction task.%

\textbf{DKT.}
We apply a simple modification to the DKT method \cite{dkt} to extend it to i) predict options instead of response correctness and ii) handle questions that are tagged with multiple subjects (the original DKT method assumes that each question is tagged with a single subject). 
We use
\[\bx_t = [\CE_q(q_t^i) \oplus \CE_o(c_t^i) \oplus \CE_s(\{s_{t}^i\}) \oplus \CE_o(a_t^i) \oplus \CE_l(r_t^i)]\]
as the input to the DKT LSTM input module.
The student's hidden knowledge states are computed using the LSTM model as $\bh_{t+1}= \text{LSTM}(\bh_t, \bx_t)$.
The predictive probabilities of selecting each option are computed using
\[\bz_t^i = f([\bh_t^i \oplus \CE_q(q_t^i) \oplus \CE_o(c_t^i) \oplus \CE_s(\{s_t^i\}) ])\in \BR^4,\, \hat{y}_t^i = \argmax_{o \in \{A,B,C,D\}} [\bz_t^i]_{o}.\] 

\sloppy
\textbf{DKVMN.}
Instead of using LSTMs to model latent knowledge state transitions, the dynamic key-value memory network (DKVMN) method uses a key-value memory network to retrieve and update knowledge at every time step using an external memory module as $\bh_{t+1}= \text{MemoryModule}(\bh_t, \bx_t)$; refer to \cite{dkvmn} for details. We use the same input and output structure for the DKVMN memory module as that for DKT. 

\textbf{AKT.}
We also adapt AKT, an attention network-based, state-of-the-art KT method for the option prediction task. AKT computes a query, a key, and a value vector for each time step, and then uses the similarity between the query and key vectors at different time steps to attend to questions in the past and use their corresponding value vectors to retrieve acquired knowledge in the past. 
We compute the query, key, and value vectors as $\mathbf{q}_t =\BW^{Q}\mathbf{n}$, $\mathbf{k}_t =\BW^{K}\mathbf{n}$, and $\mathbf{v}_t =\BW^{V}[\CE_l(r_t^i) \oplus \CE_q(q_t^i) \oplus \CE_o(y_t^i) \oplus \CE_o(c_t^i)]$ respectively, 
where $\BW^{Q}$, $\BW^{K}$, and $\BW^{V}$ are the query, key, and value projection matrices and $\mathbf{n}=[\CE_q(q_t^i) \oplus \CE_s ( \{s_{t}^i\} )\oplus \CE_o(c_t^i)]$. 
The retrieved latent knowledge state is then computed as
$\bh_t = g \Big(\sum_{\tau<t} \alpha_{t,\tau} \mathbf{v}_{\tau}\Big),$
where $g$ is another feedforward network and $\alpha_{t,\tau}$ is the normalized attention score between the query at the current time step $t$ and the key at a past time step $\tau$.
For AKT, we employ the exponential decay module to compute the attention scores \cite{akt} and then compute the output using the attention-weighted value $\bh_t^i$ and a fully connected network $f(\cdot)$. 

\section{Experiments}
\textbf{Experimental Setup.}
In addition to the option prediction task, we also evaluate all methods under the standard, binary-valued response correctness prediction task. We do not need to use a separate set of methods; instead, we can simply replace the final output layer of the option predictor module ($f: \cdot \rightarrow \BR^4$) with an output layer that consists of a single node ($f: \cdot \rightarrow \BR^1$) for all OT methods; the resulting loss function corresponds to standard binary cross entropy loss. 
For option prediction, we use both accuracy and macro $F_1$ score as evaluation metrics. 
For correctness prediction, we use accuracy as the only evaluation metric which aligns with the option prediction task. 
We compute the $F_1$ score for each question-option pair individually and average across all such pairs. This metric treats every option in every question equally, thus magnifying the impact of options that are rarely selected.
For reference, 
on the Eedi and EdNet datasets, the selection probabilities across options for an average question (from most frequent to least frequent) are 57\%, 25\%, 11\%, 7\% and 66\%, 20\%, 10\%, 4\%, respectively. 
For option prediction, a random classifier has an average macro $F_1$ score and an accuracy score of $0.25$ on both of these datasets, while a majority class classifier has an average macro $F_1$ score (accuracy) of $0.184$ (57\%) and $0.205$ (66\%) on the Eedi and EdNet datasets, respectively.

 \begin{table}[t]\centering
     \scalebox{.84}{
     \begin{tabular}{c | cc | cc|  cc }\toprule
         & \multicolumn{4}{c|}{Option Prediction}&  \multicolumn{2}{c}{Correctness Prediction}\\
         \cline{2-7}
         & \multicolumn{2}{c|}{Accuracy}  & \multicolumn{2}{c|}{Average Macro $F_1$ Score} & \multicolumn{2}{c}{Accuracy} \\
         \cline{2-7}
         Model & Eedi & EdNet  & Eedi & EdNet & Eedi & EdNet \\
         \midrule
NCF & 64.75$ \pm $0.02 & 67.24$ \pm $0.01 & 0.2824 $\pm$ 0.002 &0.2552 $\pm$ 0.001 & 72.6$ \pm $0.03 & 71.49$ \pm $0.01  \\
PO-BiDKT & 65.87$ \pm $0.01 &{\bf 69.42$ \pm $0.01} & {\bf 0.3283 $\pm$ 0.001}& {\bf 0.3260 $\pm$ 0.001 }&  75.18$ \pm $0.01 & {\bf 75.21$ \pm $0.02}  \\
BiGIKT & {\bf 66.16$ \pm $0.02} & 69.29$ \pm $0.02 &0.3261 $\pm$ 0.001 & 0.3168 $\pm$ 0.001 & {\bf 75.62$ \pm $0.02} & 75.07$ \pm $0.01 \\
\midrule
DKT & 65.95$ \pm $0.44 & 68.03$ \pm $0.09 &0.313$\pm$ 0.008 & 0.2887 $\pm$ 0.005 & 74.7$ \pm $0.34 & 73.19$ \pm $0.06  \\
DKVMN &{\bf 66.03$ \pm $0.49} & 68.01$ \pm $0.1 &{\bf 0.3152 $\pm$ 0.007} &0.2842 $\pm$ 0.005 & {\bf 74.75$ \pm $0.3} & 73.02$ \pm $0.06  \\
AKT & 65.91$ \pm $0.47 & {\bf 68.44$ \pm $0.09 }& 0.3139 $\pm$ 0.007 & {\bf 0.3062 $\pm$ 0.004} & 74.65$ \pm $0.31 & {\bf 73.6$ \pm $0.06}  \\
\toprule
     \end{tabular}
     }
     \caption{Performance of all methods under the CF (top half) and KT (bottom half) setups on both datasets. Best results are in \bf{bold}.}
     \label{tab:cf}
   \vspace{-0.5cm}
 \end{table}

\textbf{Training and Testing.}
We perform standard $k$-fold cross-validation (with $k = 5$) for all methods on both datasets. Under the CF setup, on average 20\% of the time steps (for each student) are randomly chosen as the held out test set, 20\% of time steps are randomly chosen as the validation set, and the other 60\% are chosen as the training set to train all methods. %
Under the KT setup, all time steps for a randomly chosen 20\% of students are used as the test set,
and the validation and training sets are constructed similarly. %

\textbf{Network Architectures and Hyper-parameters.} Since the datasets are large, we do minimal hyper-parameter tuning and set most of the values to their default values for all the methods; exploratory experiments found that evaluation results are robust across most parameter values. 
We set the question, subject, option, response embedding dimension for all methods to $d \in \{32,64\}$ for the CF setup and $d \in \{64,128\}$ for the KT setup. 
We use the Adam optimizer \cite{adam} to train all models with a batch size of $64$ students to ensure that an entire batch can fit into the memory of our machine (equipped with one NVIDIA Titan X GPU). For all methods, we set the learning rate to $10^{-4}/10^{-3}$ for the Eedi/EdNet dataset and run all the methods for $200$ epochs and perform early stopping based on the loss on the validation set. 
We set the latent knowledge state ($\bh_t$) dimension to $256/512$ for all methods under CF/KT setup. For NCF, we select the user embedding dimension as $d = 256$.

\textbf{Results and Discussion.} Table~\ref{tab:cf} lists the performance of all OT methods for both the CF and KT setups for both the option prediction and correctness prediction tasks, on both datasets; we report the averages as well as the standard deviations across the five folds. We observe a significant dropoff ($\sim 10\%$) in the accuracy metric on the option prediction task compared to the correctness prediction task, which is as expected since there are four categories to predict $(A,B,C,D)$ instead of two categories (correct/incorrect). As a result of this difference, the correctness prediction task can be seen as a sub-task in the option prediction task by computing the probability a student selects the correct option. The performance of different methods are also quite consistent across all cases.

We observe that recurrent neural network-based methods such as PO-BiDKT perform significantly better than NCF in all cases. 
This observation suggests that even in a CF setup for model evaluation, methods that take the evolving nature of student knowledge into account are still more effective than popular CF methods that do not account for these temporal dynamics. 
Overall, we observe that the performance gains on the option prediction task provided by complex model architectures are marginal.  
This observation suggests that more work needs to be done on the option prediction task to understand the dynamics behind students' decisions to select a specific incorrect option, which motivates our exploration in Section~\ref{sec:errormode}. 
In the KT setup, we observe that DKVMN performs best on the Eedi dataset while AKT performs best on the EdNet dataset. This observation suggests that complex neural network architectures such as attention modules are more beneficial when a large amount of training data is available. 

In both setups, we observe that the $F_1$ scores are low for all methods; despite clearly not simply predicting the most frequent option, the performance of these methods leaves significant room for improvement due to class imbalance. 
Possible approaches to improve prediction accuracy for options that are rarely selected include oversampling them \cite{smote}; however, since a student's responses to different questions are not independent data points, how these methods can be applied to the option tracing task is not immediately clear.

\vspace{-0.25cm}
\subsection{Clustering Incorrect Options}
\label{sec:errormode}

To qualitatively evaluate our option tracing methods, we attempt to group incorrect options across multiple questions into clusters and examine whether question-option pairs in the same cluster correspond to the same underlying error. To this end, we train a modified version of PO-BiDKT on the Eedi dataset \cite{neuripschal}; we learn an embedding module $\CE_{q,o}(q,o):q\times o\rightarrow \BR^d$ for each question-option pair. Then, we compute the option selection probabilities using the latent knowledge state $h_t^i$ and the question-option pair embeddings as $p(o|q_t^i) = \frac{f(h_t^i)^T \CE_{q,o}(q_y^i,o) }{\sum_{o'}f(h_t^i)^T \CE_{q,o}(q_y^i,o')}$.
This modification suffers a small drop in predictive performance but encodes information in the question-option pair embeddings for us to cluster them and search for common student errors.

 \begin{table}[t]\centering
     \scalebox{.95}{
     \begin{tabular}{c | c |c  }\toprule
Metric & Adjusted Rand Index & Fowlkes-Mallows index \\
\midrule
Score & 0.372 &  0.455 \\
\toprule
     \end{tabular}
     }
     \caption{Incorrect option clustering quality for a subset of questions in the Eedi dataset using errors labeled by a domain expert. $1$ in both metrics indicates perfect clustering.}
     \label{tab:cluster2}
    \vspace{-0.6cm}
 \end{table}
 
We selected all incorrect options ($31\!\times\! 3\!=\!63$) in questions on subject $33$ (BIDMAS) where question images are released on the smaller subset of the Eedi dataset; see \cite{neuripschal} for details. A domain expert manually labeled each option based on which error likely resulted in the student selecting it, resulting in a total of $14$ high-level errors (errors that cannot be named are excluded), each corresponding to multiple options across different questions; further splitting them into finer-grained errors results in clusters that are not meaningful. We perform k-means clustering \cite{kmeans} on the learned question-option pair embeddings and compare them to the ``ground truth'' option clusters provided by the expert.

Due to spatial constraints, we only report quantitative results on clustering quality using two commonly used metrics: The adjusted Rand index \cite{ari-score} and the Fowlkes-Mallows index \cite{fmi-score}; the former has a range of $[-1,1]$ while the latter has a range of $[0,1]$, with $1$ corresponding to perfect clustering. %

Table~\ref{tab:cluster2} lists these metrics on the learned question-option pair embeddings based on the ground truth expert labeling. Overall, the clustering performance is acceptable but not excellent. We observe that some errors such as ``sign error in calculation involving negative numbers'' have relatively easy-to-identify corresponding option clusters ($5$ out of $8$ options labeled by the expert as corresponding to that error are put into the same cluster). On the other hand, some options such as $69D$ and $293C$ (the left half of Figure~\ref{fig:misc}) correspond to the same error but are not grouped into the same cluster. One possible explanation is that students may not consistently demonstrate an error, as found in prior research \cite{kurt}; among students who selected $69D$, only $51\%$ selected $293C$ while $34\%$ of them selected the correct option, $293B$. Therefore, further work is required to study whether more robust KT methods and clustering algorithms can identify error clusters more effectively. Nevertheless, our approach produces a starting point to reduce the effort for domain experts to manually label errors and provides them a way to do it under data-driven support.

\section{Conclusions and Future Work}
Analyzing the exact options students select across multiple choice questions has the potential to uncover their error modes and help teachers to provide targeted feedback to improve learning outcomes. 
In this paper, we proposed a set of methods to extend common knowledge tracing methods that analyze only the correctness of students' responses to questions to analyze the exact options they select on multiple choice questions. We validated these methods with quantitative experiments on two large-scale datasets in terms of their ability to predict the options students select on each question and qualitative experiments in terms of clustering incorrect options according to underlying errors.
There are many avenues for future work. First, we need to develop methods that are aware of the evolving nature of student errors. One possible approach is to develop methods that can explicitly account for the recurrence of past errors, such as using a neural copy mechanism \cite{copy}; these methods may help us track students' progress in correcting their errors. Second, low $F_1$ scores for the option prediction task suggest that it is much more challenging than the typical correctness prediction task in knowledge tracing literature and thus deserves more attention.

\bibliographystyle{splncs04}
\bibliography{references}

\end{document}